\documentclass[lettersize,journal]{IEEEtran}
 
\usepackage[dvipsnames]{xcolor}
\usepackage{caption} 
\usepackage{multirow} % for cmd 'multirow', 'multicolumn'
\usepackage{pdfpages}
\usepackage{amsmath,amsfonts}
\usepackage{algorithmic}
\usepackage{array}
\usepackage[caption=false,font=normalsize,labelfont=sf,textfont=sf]{subfig}
\usepackage{textcomp}
\usepackage{stfloats}
\usepackage{url}
\usepackage{verbatim}
\usepackage{graphicx}
\usepackage{algorithm}
\usepackage{booktabs}
\usepackage{algorithmic}
\usepackage{booktabs}
\usepackage{stfloats}
 
\usepackage{xcolor}
\def\BibTeX{{\rm B\kern-.05em{\sc i\kern-.025em b}\kern-.08em
    T\kern-.1667em\lower.7ex\hbox{E}\kern-.125emX}}
\usepackage{balance}
\begin{document}
\title{AbHE: All Attention-based Homography Estimation}
\author{Mingxiao Huo, Zhihao Zhang, Xinyang Ren, Xianqiang Yang

\thanks{Mingxiao Huo, Xinyang Ren and Xianqiang Yang are with the Research Institute of Intelligent
Control and Systems, Harbin Institute of Technology, Harbin 150001, China
(e-mail: 1190600119@hit.edu.cn; 16B904014@stu.hit.edu.cn; xianqiangyang@hit.edu.cn).}
\thanks{Zhihao Zhang is with the College of Electrical Engineering
and Control Science, Nanjing Tech University, Nanjing 211816, China
(e-mail: zhihaozhang94@njtech.edu.cn).}
\thanks{(Corresponding Author: Xianqiang Yang)} }

\maketitle

\begin{abstract}
Homography estimation is a basic computer vision task, which aims to obtain the transformation from multi-view images for image alignment. 
Unsupervised learning homography estimation trains a convolution neural network for feature extraction and transformation matrix regression. While the state-of-the-art homography method is based on convolution neural networks, few work focuses on transformer which shows superiority in high-level vision tasks. In this paper, we propose a strong-baseline model based on the Swin Transformer, which combines convolution neural network for local features and transformer module for global features. Moreover, a cross non-local layer is introduced to search the matched features within the feature maps coarsely. In the homography regression stage, we adopt an attention layer for the channels of correlation volume, which can drop out some weak correlation feature points. The experiment shows that in 8 Degree-of-Freedoms(DOFs) homography estimation our method outperforms the state-of-the-art method.
\end{abstract}


\section{Introduction}
\IEEEPARstart {T}{he} traditional homography estimation, which is of vital role in image alignment~\cite{zhang2022image,zhang2021natural}, always involves feature points extraction, feature match algorithm. In the first step, the traditional homography estimation always adopts some classical feature extractor. like ORB~\cite{rublee2011orb}, SURF~\cite{bay2006surf}, and SIFT~\cite{lowe2004distinctive}. In the feature match part, the RANSAC~\cite{fischler1981random} algorithm is widely applied to estimate an accurate homography, through the search for the most matching feature point pairs. The 2D homography estimation is a crucial part in the monocular SLAM system,which relies on the feature extraction, feature matching and the transformation among different frames. Because homography is a transformation between two images caused by the rotation of the camera center, it can be used in such scenes in the SLAM system: the camera transformation contains pure rotation, or the camera is far from the objects. In the ORB-SLAM~\cite{mur2015orb}, they combine the homography estimation with the matrix estimation. 

However, the traditional homography estimation methods are seriously affected by the density of the feature points, especially in a monotonous scene with sparse feature points. For the superiority of deep neural network, supervised learning method is introduced to estimate homography~\cite{detone2016deep}, in this method, the transformation between the source images and the target images is predicted directly by a neural network, of which an objective optimization to minimize the gap between the labeled homography and the homography generated by neural network. However, it requires much more labors for the homography annotations and is constrained in synthetic datasets. Therefore, the unsupervised learning method is becoming more acceptable in homography estimation.
\begin{figure}[H]
\centering  
\includegraphics[width=0.53\textwidth]{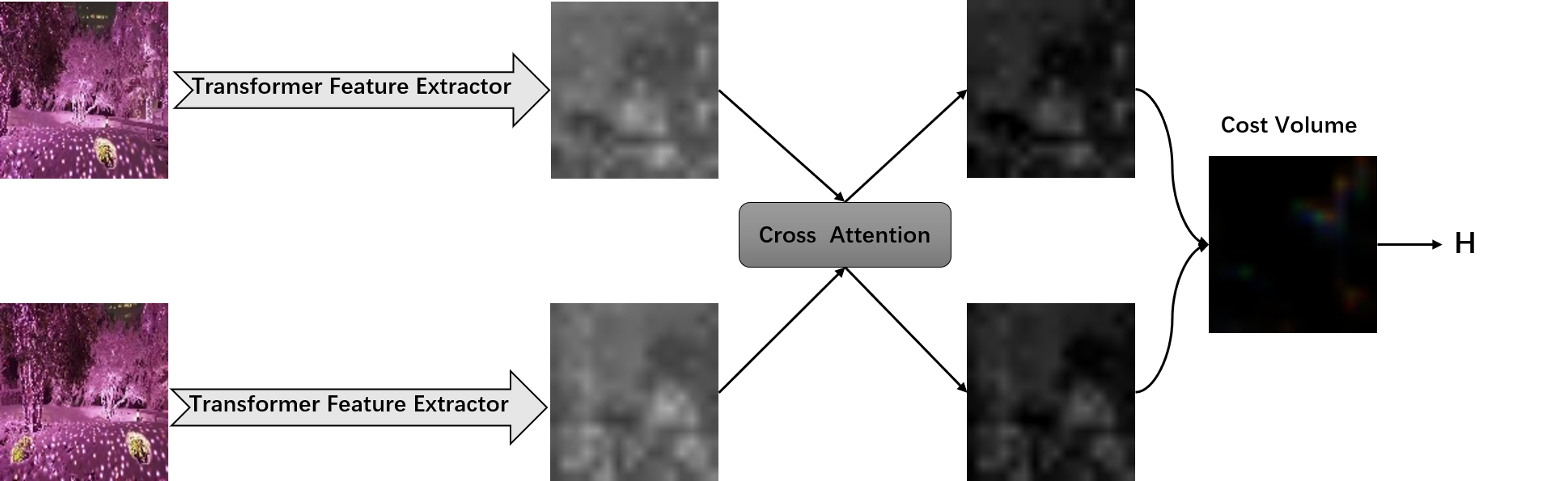}  
\caption{The visualization of the forward process for our method. First, we extract the feature map of the source and target images by transformer style backbone. Second, through a cross attention layer, we get an initial filtration for the candidate feature points. As the figure shows, after the cross attention, some feature points become more obvious than the points around.  Third, we calculate the cost volume for two feature maps, which can store the position information about the matched keypoint pairs. At last, we regress the final homography with a channel attention.}  
\label{Fig.main2}  
\end{figure}
  A classical unsupervised learning method~\cite{nguyen2018unsupervised} dexterously optimizes the pixel distance between the pair of the source and target images, which no longer requires the labeled data. Then, to focus on more keypoints, content aware unsupervised method~\cite{zhang2020content} introduces an attention map to reject invalid regions. However, these methods can only be appropriate for a large baseline scene, where more content region features can be
  provided. To tackle this problem, a view free pipline~\cite{nie2020view} is realized by adopting a coarse-to-fine strategy. In this paper, we inherit this strategy for its application in more flexible scenes. Simultaneously, we need to take advantages of transformer in computer vision, which can cover a large range to collect feature information for a feature point. Thus, we unlock the Swin Transformer structure~\cite{liu2021swin}in this paper, which hierarchically extracts transformer-style feature maps from coarse to fine.
  
  Recently, transformer structure has been widely applied in vision tasks as an alternative to CNN~\cite{carion2020end,dosovitskiy2020image,liang2021swinir}, because it can capture a global context interaction by a self-attention mechanism. But the transformer with a fixed patch size, like vision transformer~\cite{dosovitskiy2020image}, has three main drawbacks for  homography estimation. First, it can bring a heavy computational burden for the interaction of all the global patches. Second, it cannot generate a feature pyramid structure, which has been proved effective in large-baseline scene. Third, it lacks the locality inductive biases, which can be vital for forecasting the similarity regions between source and target image pairs. Swin Transformer can alleviate these problems by a hierarchical structure, which can generate local feature maps in various sizes. Also, in this paper, we adopt cnn as the first layer, which is regarded as a useful way to extract some shallow features, like colors, textures.
  
  In the traditional transformer structure~\cite{vaswani2017attention}, they design an attention mechanism, which gathers the key and query from encoder and decoder part separately. Then, the key information from encoder will better align with the query information. Thus, we rethink this idea for the homography estimation task, which requires the alignment between the corresponding key points in source and target images. In this paper, we respectively regard the source and target images as key and query, to search the matching feature points globally. For a large range key points search, we adopt a non-local self-attention mechanism. 
  
  RANSAC algorithm~\cite{fischler1981random} is always an indispensable process in homography estimation, which can reject many outlier points and focus on more feature points which provide higher precision. In the regression stage, we want to put forward the similar idea, which can relieve the filtrate burden for the regression neural network. Many plug-and-play attention modules~\cite{hu2018squeeze,woo2018cbam} have been widely used in computer vision, which can be also used in this task to filter the interior points. We first get a correlation volume for the feature maps, whose every channel represents the correlation matrix between one point in source image and all points in target image. Then, a channel attention layer is adopted for the correlation volume, which can discard some weak correlation points, like the RANSAC process.
  
  We demonstrate the effectiveness of our pipline and proposed new components by comparison experiments and ablation studies. The experiment result shows that our method outperforms the state-of-the-arts on the public dataset quantitatively. Also, the comparison experiment shows our method is competitive in quality. The main contributions are summarized as follows:
  \begin{itemize}
    \item We propose a swin-transformer based homography estimation model, with abundant parameters and large model size, which can  provide a  long-range dependency modelling by transformer and short-range dependency modelling by cnn.
    \item Considering the search for the matching points globally and advancedly, we propose a cross non-local attention mechanism, thus achieving a pre-alignment effect for the corresponding feature representations.
    \item A channel attention module is applied for the correlation volume, which drops out some weak correlation point pairs and achieves the effects similar with RANSAC process in traditional method.
\end{itemize}
\section{Related Work}
In this section, we will briefly introduce some homography estimation methods and some basic ideas of the attention mechanism in the computer vision tasks.
\subsection{Feature-based Homography Estimation}
Before the boom of the deep neural network, the homography estimation mostly adopts the machine learning methods for the feature extraction, and the matching algorithm. In the dual-homography warping (DHW) algorithm~\cite{gao2011constructing}, after the feature extraction, they cluster the feature points and separate them into ground plane and distant plane. Because in many cases, different parts of the scenes will have totally different homography estimation due to the distance between the camera and the objects, this method is a breakthrough compared with traditional single part feature matching~\cite{hartley2003multiple}, ~\cite{brown2007automatic}. In ~\cite{zaragoza2013projective}, they partitioned the whole image into many grids, and computed the homography for each grids locally, which is used to improve the accuracy of the image alignment.

\subsection{Learning-based Homography Estimation}
The main backbone for learning-based homography estimation is firstly developed by ~\cite{detone2016deep}. In this paper, they firstly adopt a convolution neural network(CNN) to regress the displacement of the four coordinates and transform the displacement to homography by Direct Linear Transformation(DLT). The displacement of the coordinates is  more valid and concrete than the homography matrix for the deep neural network to learn. Based on this method,  ~\cite{nguyen2018unsupervised} develops unsupervised homography estimation, removing the large burden for labelling. This paper designs the loss function by measuring the content difference  between reference and warp images without the requirement for generating the labels of the true homography. In ~\cite{chang2017clkn} and ~\cite{erlik2017homography}, they all employ a hierarchical neural network to estimate the homography accurately step by step. Following the estimation from coarse to fine strategy, the ~\cite{nie2020learning} adopts the feature pyramid structure to deal with the large-baseline scene problems. For the small-baseline scene, the content aware homography estimation ~\cite{liu2022content} which employs an attention content mask to reject the invalid feature points effectively, and shows the superiority. In the ~\cite{koguciuk2021perceptual}, they employs a frozen loss network to calculate the content loss, which is more stable for representing the content in a deeper level of the neural network just as the style transfer work ~\cite{gatys2016image}. Back to the original idea about regressing the four corners' displacement, the ~\cite{Ye_2021_ICCV} changes the idea, by defining a homography flow representation, which is computed by the regression results, which represent 8 flow bases.
\subsection{Attention in Computer Vision}
Attention mechanism is firstly developed in natural language processing as a method of modeling the interaction of a sequence, like RNN ~\cite{medsker2001recurrent}, LSTM ~\cite{hochreiter1997long}. The original attention mechanism idea in computer vision is not the same style as its in natural language processing. The inchoate attention applied in computer vision mostly about focuses on some areas, or some channels by learning an attention mask, like SENet ~\cite{hu2018squeeze}, Cbam ~\cite{woo2018cbam} and DANet ~\cite{fu2019dual}, they achieve a considerable result in many computer vision tasks, like image classification, object detection and scene segmentation. These methods can also be seen as a method of building sequence model, because focusing on different parts or channels can be seen as paying attention to different tokens. In the action recognition, a non-local neural network ~\cite{wang2018non} is introduced to build the interaction about all the pixels and focus on the area which is highly related with the action. As a global model, it covers a lot of memory, so it is mostly used for the feature maps with a smaller size. As the transformer structure ~\cite{vaswani2017attention} becomes more and more competitive in natural language processing, the VIT ~\cite{dosovitskiy2020image} partitions one image into different tokens, and then feeds these tokens into transformer block. With the advantage of long-term modelling, the transformer transcends the cnn in many areas of computer vision, like video understanding ~\cite{arnab2021vivit} and multi-model tasks ~\cite{bain2021frozen}.

However, the vision transformer requires a heavy calculation burden, because of the global interaction. Considering such a limitation, the swin transformer ~\cite{liu2021swin} is proposed to calculate the attention just in a window area. To avoid the lack of global modelling, it also employs a shifted window mechanism to interact the pixels from different windows. Also, the swin transformer applies a hierarchical structure, which can highly reduce the computational complexity, which is mostly decided by the image size. The swin transformer further proves the superiority of the attention mechanism in computer vision tasks, which achieves competitive results in image restoration ~\cite{liang2021swinir}, video understanding ~\cite{liu2022video} and image segmentation ~\cite{lin2022ds}.
\begin{figure*}
\centering  
\includegraphics[width=1\textwidth]{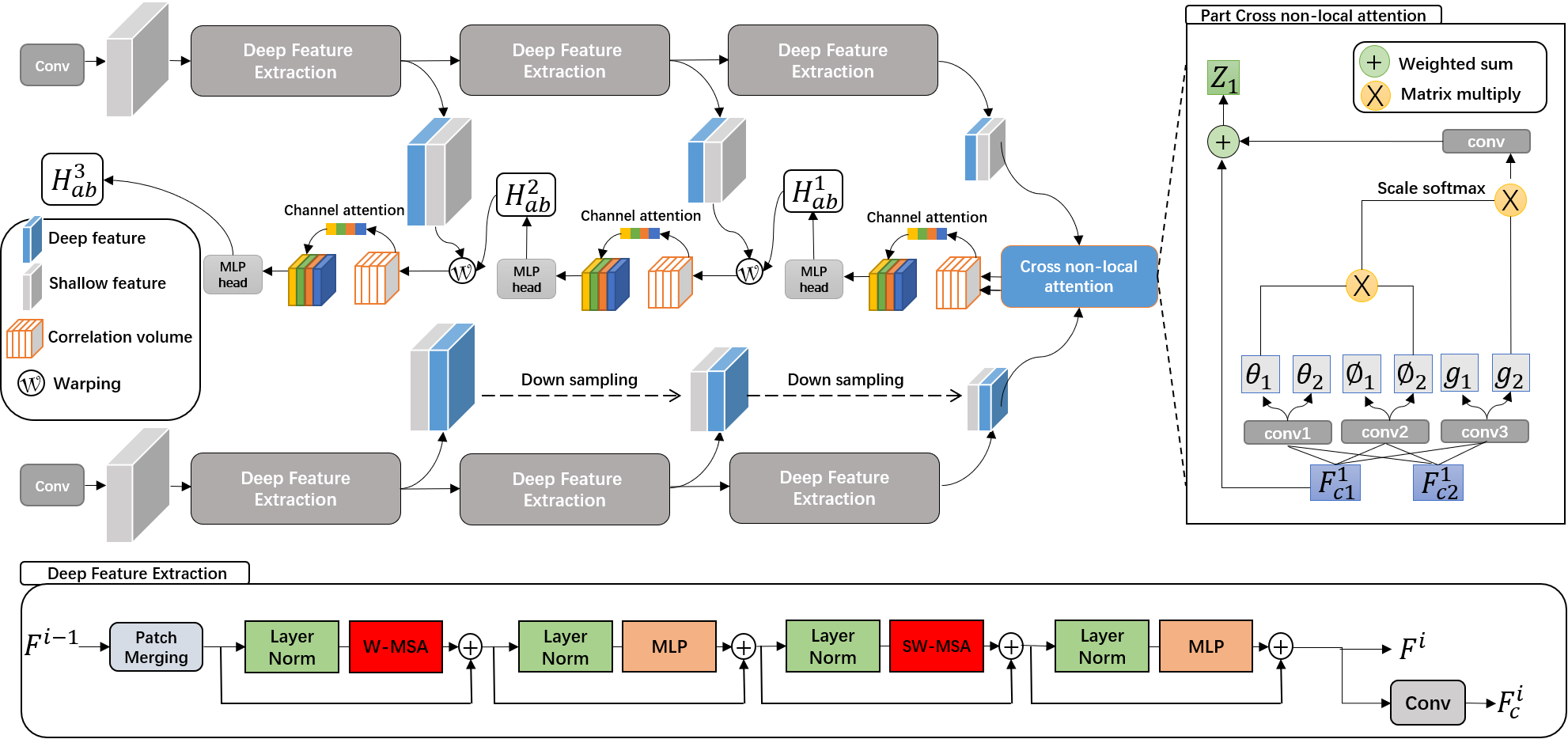}  
\caption{The architecture of our proposed model for homography estimation. The network mainly consists of shallow feature extraction, deep feature extraction, and cross non-local attention layer. Noticeably, half part of the non-local attention is illustrated in the figure, the other half part is symmetrical with this part. The whole network inputs are images from two views, and the output is the homography transformation between the two images.}  
\label{Fig.main2}  
\end{figure*}
\section{Method}
\subsection{Motivation}
For a long time, the unsupervised learning method for homography estimation mainly bases on CNN method~\cite{nie2021depth,nie2021unsupervised,ye2021motion}, and achieves relatively competitive results, even compared with supervised learning method. However,it is rare to adopt a transformer architecture in the previous work, ~\cite{hong2022unsupervised} utilizes a transformer structure only in the regression stage. When we are rethinking the homography estimation task, a feature point which gathers the global information can be easier to be distinguished from the similar adjacent points than the local modelling methods, like cnn. Thus, considering the strong global modelling ability for the transformer, we unlock a Swin Transformer backbone for feature extraction stage in this paper. Additionally, the Swin Transformer, which can provide a multi-level feature pyramid, fits the large-baseline homography estimation task naturally. Also, another advantage for attention mechanism is that it can realize the interaction between the vision regions, it always lacks a valid connection modelling between source and target feature maps in the previous work, so we design a cross non-local attention for the feature maps from different views.

 As shown in Fig.1, two original images $I_a$ and $I_b$ are input in this our structure, where $I_a,I_b\in\mathbb{R}^{H\times W\times C_{in}}$ (H, W, $C_{in}$, are height, width and the input channel, respectively). In the feature extraction process, both the feature extractors for source image$I_a$ and target image$I_b$ are the same structure and share the same weight, so we adopt $F_i$ as every feature map from both the feature extractors. Then we extract a shallow feature first by a CNN encoder, which is an outstanding feature extractor for some basic representations. After that, we feed the shallow feature to the deep feature extraction part, by a Swin Transformer~\cite{liu2021swin} based model. For the patch merging operation as the start of Swin Transformer block, the $i_{th}$ feature map will resize from original size to half of the size, and the channel number doubles, from $H_i\times W_i\times C_i$   to   $\frac{H_i}{2}\times \frac{W_i}{2}\times {2C}_i$. After the deep extraction, we resize the shallow feature to the same size as the deep feature, and concatenate them together as our every component of the feature pyramid as:
\begin{equation}
 F_i=concatenate(F_{SFi},F_{DFi}),
\end{equation}
where $F_{SFi}$ denotes the resized shallow feature map, which is the same size as the deep feature map, and $F_{DFi}$ is the deep feature map extracted by the $i_th$ Swin Transformer block.

After getting the two feature maps $F_a^i$,$F_b^i$, (i represents the feature map from the $i_th$ deep feature extraction block, a and b represent them from the source and target image, respectively), for the deepest feature maps, we design a cross non-local attention to align the feature maps in advance, for the other layers' feature maps, we calculate the correlation volume for the two $H_i\times W_i\times C_i$ feature maps. Following this, we feed the correlation volume to a regression network, consists of the channel attention and MLP layers. The former is designed to drop out some outlier feature points, which is realized by a channel attention layer. The latter is a regression network ,which is applied to regress the homography transformation between the two $i_{th}$-layer feature maps from a to b, denoted as $H_{ab}^i$. The homography transformation is predicted by a differentiable direct linear transformation(DLT) layer, with the input of 8 corner offsets, as~\cite{nguyen2018unsupervised}. Then, we warp the $(i-1)_{th}$ source feature map by the predicted transformation $H_{ab}^i$ , as the input of next regression layer, which follows the coarse-to-fine strategy, this process can be formulated as:
\begin{equation}
 H_{ab}^{i-1}= \mathcal{R}(F_b^i\otimes\mathcal{W}(F_a^i, H_{ab}^i)),
\end{equation}
where $\mathcal{R}$ denotes the regression layer, $\otimes$ denotes a matrix multiplication to get a correlation volume, and $\mathcal{W}$ denotes the warp operation for the feature map. Thus, we can predict the homography transformation in a cascade, and get an accurate estimation for the transformation.

For the whole network structure, the core feature extraction is implied by the transformer for deeper features and CNN for shallow features. For the deepest features, we design a cross non-local for a pre-alignment. Simultaneously,for the other layers, we warp the feature map by the transformation matrix predicted by the last layer, both achieves a pre-alignment effect. In the regression process, we adopt channel attention for the correlation layer, and regresses the 8 corner offsets in the end.
\subsection{Feature Extraction}
We first employ two $3\times3$ convolution layers, which is stable for early visual processing~\cite{xiao2021early}, and then feed the shallow feature maps to a Swin Transformer block.

\textbf{Swin Transformer block} consists of layer normalization~\cite{ba2016layer}, windows multi-head self-attention layer, MLP layers and shift window operations. The main difference between Swin Transformer, and basical vision transformer~\cite{dosovitskiy2020image} is that Swin Transformer adopts a local attention in a window, which is greatly decease the computation complexity, and its shift window mechanism allows the interaction for embedding tokens from different windows. After the patch merging, the feature map is resized to $\frac{H_i}{2}\times \frac{W_i}{2}\times {2C}_i$, then it is partitioned to M windows, in this paper M is set to be 4. $\frac{H_i\times W_i}{4M^2}$ windows, with the size of $M^2$ are input into a standard transformer layer for a calculation of the local attention. For each local feature window W, we calculate the query matrix Q, the key matrix K and the value matrix V as follows:
\begin{equation}
  Q=WP_Q,  K=WP_K,  V=WP_V,
\end{equation}
where $P_Q$, $P_K$, $P_V$, are the query, key and value matrices for the window features. Then, after introducing a positional encoding b, attention matrix can be computed as:
\begin{equation}
Attention(Q,K,V)=SoftMax(QK^T/\sqrt{d}+b)V,  
\end{equation}

Then, after another layer normalization stage, the attention tokens are input in a  MLP layer with two fully connected layers and a RELU activation layer, A residual structure~\cite{he2016deep} is employed in the Swin Transformer block as a connection for different levels' representations. Because the previous structure is fixed in the size for the local attention mechanism, the Swin Transformer introduces a shifted window partitioning with each window shifting $\frac{M}{2}$ to interact the features in one local window with the features in other windows.

After each deep feature extraction by transformer, we adopts a CNN layer as Fig. 1 shown, which can keep the same feature style with the shallow feature extraction, and can introduce inductive bias for the network. And the feature directly from Swin Transformer block $F^{i'}$ is input in the next Swin Transformer block to keep the consistency for the features.

\subsection{Cross Non-local Attention}
Attention can be used to compute the relationship between two regions, which is highly related with homography estimation problems to search the similar regions between source and target images. Therefore, we introduce an attention mechanism for the feature maps from the two input images. Because there is no pre-warp operation for the last feature map, we design this cross non-local layer only for the deepest feature map, which leads to a heavy memory burden for GPU.

As shown in algorithm 1, the third feature maps $F_a^3$ and $F_b^3$ are projected into a low-dimension embedding space by three different $1\times 1$ convolution, \emph{i.e.}, key projector, query projector and value projector. The same projector for source feature map and target feature map both share the same weight, to guarantee the feature style invariant in the low-dimension space. Then, we flat the feature map in the $H\times W$ dimension, so we reshape the feature map from $H\times W\times C$ to $HW\times C$. After that, we imply cross multiplication in the channel dimension between key projection and query projection from different maps, which can be expressed as:
\begin{equation}
   S_i= Q(F_i^3)^T K(F_j^3), i,j\in \{a,b\}, i\neq j,
\end{equation}
where Q denotes the query projector, and K denotes the key projector. Because of the flatten operation and channel multiplication, every single point of $S_i$ represents the similarity degree between all the feature points in the i feature map and all the feature points in the other feature map. For $i\neq j$, the two feature maps  mutually calculate the similarity matrix to search a high-similarity position. Then, we adopt a scale softmax for every column of the similarity matrix, following~\cite{nie2021depth}, which can promote a strong similarity, and restrain a weak similarity, by adjusting a temperature hyperparameter k. With the increase of k, the effect will become more significant. In this paper, k is set to be 10.

Getting the weighted similarity matrix, we do another multiplication for the value projection and similarity matrix, then we can get a reconstructed feature map, which can be shown as:

\begin{algorithm}[tb]
    \caption{Pseudocode of Cross Non-local Attention}
    \label{alg:algorithm}
    \textbf{Input}: $F_a^3$, $F_b^3$\\
    \textbf{Parameter}:\\
    k: temperature parameter for scale softmax\\
    $\lambda$: weighted sum parameter\\
    \textbf{Output}: $Z_a$, $Z_b$\\
    %[1] enables line numbers
    ${\theta}_{1}=conv1(F_a^3)$ \textcolor{SpringGreen}{\# query projector} \\
    ${\theta}_{2}=conv1(F_b^3)$ \textcolor{SpringGreen}{\# query projector}\\
    ${\phi}_{1}=conv2(F_a^3)$ \textcolor{SpringGreen}{\# key projector}\\
    ${\phi}_{2}=conv2(F_b^3)$ \textcolor{SpringGreen}{\# key projector}\\
    ${g}_{1}=conv3(F_a^3)$ \textcolor{SpringGreen}{\# value projector}\\
    ${g}_{2}=conv3(F_b^3)$ \textcolor{SpringGreen}{\# value projector}\\
    \textcolor{SpringGreen}{\# batch matrix multiplication for query and key}\\
    $S_a=bmm({\theta}_{1}.view(B,HW, C), {\phi}_{2}.view(B,C,HW))$\\
    $S_b=bmm({\theta}_{2}.view(B,HW, C), {\phi}_{1}.view(B,C,HW))$\\
    \textcolor{SpringGreen}{\# scale softmax in the last dimension}\\
    $S_{a}\_softmax=softmax(S_a*k, -1)$\\
    $S_{b}\_softmax=softmax(S_b*k, -1)$\\
    \textcolor{SpringGreen}{\# batch matrix multiplication for f and value}\\
    $y_a=bmm(S_{a}\_softmax,g_2).view(B,H,W,C)$\\
    $y_b=bmm(S_{b}\_softmax,g_1).view(B,H,W,C)$\\
    \textcolor{SpringGreen}{\# output projector}\\
    $\omega\_y_a=conv4(y_a)$\\
    $\omega\_y_b=conv4(y_b)$\\
    \textcolor{SpringGreen}{\# weight sum}\\
    $Z_a=\lambda\cdot F_a^3+(1-\lambda)\cdot \omega\_y_a$\\
    $Z_b=\lambda\cdot F_b^3+(1-\lambda)\cdot \omega\_y_b$\\ 

\end{algorithm}

\begin{equation}
   y_i= S_{wi}^T V(F_j^3), i,j\in \{a,b\}, i\neq j,
\end{equation}
where V denotes the value projector and $S_{wi}$ denotes the weighted similarity matrix, $S_{wi}\in\mathbb{R}^{HW\times HW}$, and $V(F_j^3)\in\mathbb{R}^{HW\times C}$, thus, every $y_i$ will be reconstructed as a feature map whose every single point gathers the similar region's information of the other feature map. For example, if the $i_{th}$ point from $F_a$ should match with the $j_{th}$ point from $F_b$, then what the reconstruction for $F_b$ is to replace the $i_{th}$ point from $F_b$ with the collection of the region around the $i_{th}$ point from $F_a$. Then another convolution is employed to project $y_i$  to an embedding $\omega\_y_i$. At the end of this attention mechanism, there is a weight sum of the original feature map and the embedding $\omega\_y_i$, which can be shown as:
\begin{equation}
    Z_i=\lambda\cdot F_i^3+(1-\lambda)\cdot \omega\_y_i, i\in \{a,b\},
\end{equation}
where $Z_i$ denotes the reconstructed feature map, $\lambda$ is a hyperparameter which controls the degree of the information conserved from the original feature maps. In this paper, we set z to 0.9, which represents a high-level original information conservation.

Therefore, the output of the cross non-local attention will contain its own information and the information from the related region from the other feature map.

The cross non-local layer is a process, which contains the search for a related region, and a global fusion of the feature information from that region. Therefore, after this process, a pre-alignment is completed for both two feature maps have the feature representation of the the other's correlated region.
 
\subsection{Channel Attention for Correlation Volume}
Given two feature maps with the same size from source and target piplines,  it requires a regression network to predict the homography transformation for this layer. The regressiont net in this paper consists three convolution layers and four fully-connected layers. However, before feeding to the regression net, there should be a reasonable
connection for the two feature maps gathering from two symmetric networks. Combining the idea for searching the most similar points in traditional homography task, we naturally employs the similarity calculation to generate a correlation volume~\cite{sun2018pwc} as the connection of the two feature maps. Therefore, we first compute the correlation volume as follows:
\begin{equation}
   C_{i,j,k\times l}=\frac{(F_a^{(i,j)})^T(F_b^{(k,l)})}{|F_a^{(i,j)}||F_b^{(k,l)}|},
\end{equation}
where $C_{i,j,k\times l}$ denotes a single point in the 3d volume, $F_a^{(i,j)}$ and $F_b^{(k,l)}$ represent a feature point in the feature maps. Thus, the $m_{th}$ channel with the size of $H\times W$ represents the correlation between the $m_{th}$ feature point of the source image and the whole target image. In our 3d volume, we calculate the correlation between a single point and a $3\times 3$  area from the other map to improve the robustness for avoiding an incorrect match between extremely similar points from a not related region.

 As shown in Fig. 2., we adopt a channel maxpooling, \emph{i.e.}, a 3d maxpooling layer. Thus, we can get the largest value in every channel with the size of $H_i\times W_i\times 1$, which represents the strongest correlation between a feature point in the source image and a feature point in the target image. Then, we train a MLP layer, with three fully connected layers, which is a bottleneck structure. The MLP layer keeps the same size as the inputs size  $H_i\times W_i\times 1$, and then we multiply the output from MLP and the correlation volume in channel, getting a weighted correlation volume. The multiplication operation in channel can restore the shape of the 3d volume, and it is of great value because every map contains the positional information of the feature points.

 The weighted correlation volume takes different weights in different channels. Because of the channel maxpooling layer and the learning process of MLP layer, some weak correlation layer will be restrained strongly for a small weight allocated. We rethink the idea in traditional homography estimation, the RANSAC algorithm aims to reject some outlier points when matching the feature points, which is visual to utilize as many as possible valuable keypoints. The channel attention for the correlation volume is a similar process with the RANSAC algorithm, because it is just a learning method to drop out these outlier points by allocating an extremely low weight for these points.
\begin{figure}[H]
\centering  
\includegraphics[width=0.5\textwidth]{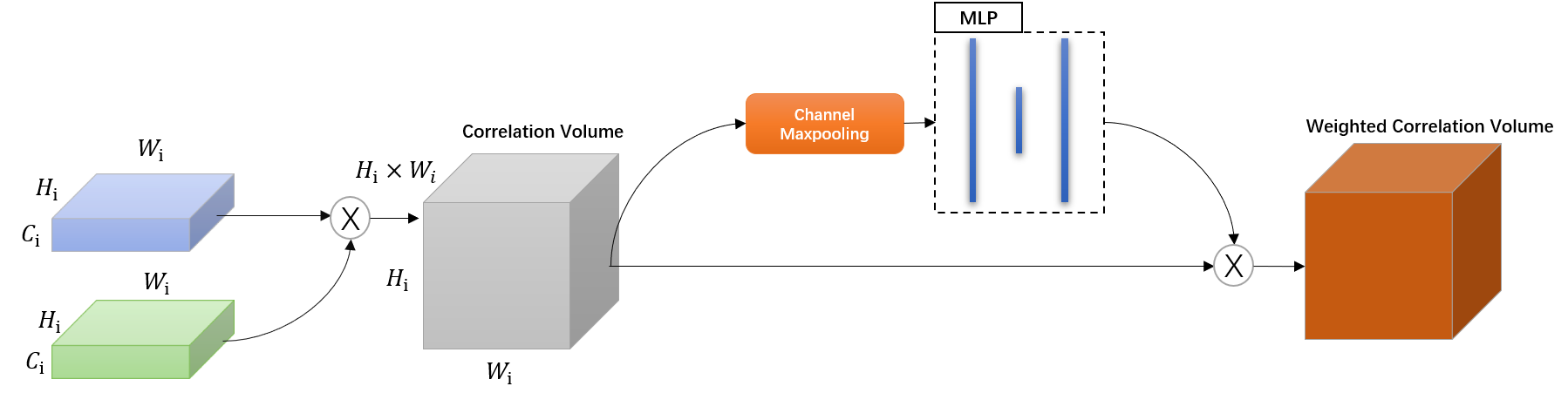}  
\caption{The process of calculating correlation volume, channel attention mechanism. The inputs are two same-sized feature maps, and the output is a weighted correlation volume.}  
\label{Fig.main2}  
\end{figure}
\subsection{Loss Function} 
In this paper, the loss function for the whole network can be divided into two parts: pixel loss, and content loss.

The pixel loss is to calculate the pixel distance after the homography transformation for the original images. Following~\cite{nie2021unsupervised}, to avoid the invalid regions, the pixel loss ablate these regions by a warped mask. The pixel objective function can be formulated as:
\begin{equation}
   L_P= \sum_{i=1}^3\omega_i||H_{ab}^{i}(E^i)\odot I_a-H_{ab}^{i}(I_b)||_1,
\end{equation}
where $H_{ab}^{i}$ denotes the homography estimated from the $i_{th}$ layer, $E^i$ denotes all-one mask with the same size as the $i_{th}$ feature map. $I_a, I_b$ are the input images, $\odot$ is the pixel-wise multiplication. $\omega_1, \omega_2, \omega_3$ are different weights for different size feature maps.

For the cross non-local layer output $Z_a, Z_b$, because its a search process for the similar regions, we propose a formulation for these two feature content outputs, which can accelerate the searching process and improve the converge effect. The content objective function can be formulated as:
\begin{equation}
   L_C=  ||H_{ab}^{3}(E^3)\odot Z_a-H_{ab}^{3}(Z_b)||_1,
\end{equation}

Taking the two part loss functions into consideration, and weight the two loss functions, the whole network's objective function can be expressed as:
\begin{equation}
   L=   \lambda_C L_C+\lambda_P L_P,
\end{equation}
where $\lambda_C$ and $\lambda_P$ represent the weights for content loss and pixel loss, respectively. With the two parts of loss functions, the model can be trained by the supervision of both high level representations (content loss) and the shallow level representations (pixel loss).

\section{Experiments}
In this section, some extensive experiments are conducted to prove the effectiveness of our method. First, we do some comparative experiments , which contain the quantitative comparison and the quality comparison. After that, we transfer our pretrained model to some real dataset and verify the strong abilty of our model in transfer learning and zero-shot inference. At last, we do some ablation experiments, combining with visualization of some parts of our neural network to prove the validity of our proposed methods.
\subsection{Dataset}
In this paper, we validate our method in the benchmark
synthetic dataset for deep homography estimation, Warped MS-COCO.The dataset contains 100,000 train images, and 10,000 test images, and is generated by MS-COCO~\cite{lin2014microsoft}. Also, we validate our method in a real dataset, UDIS-D, which is a dataset proposed in ~\cite{nie2021unsupervised} for image alignment and image stitching. The dataset contains 10,440 training images and 1,106 testing images.
\subsection{Implement Details}
 In this paper, we adopt an Adam optimizer~\cite{kingma2014adam} and employ an exponentially decaying learning rate with an initial value of $10^{-4}$. For the MS-COCO dataset the training batch size is set to 16, and the training iteration is 240,000 steps, about 38.4 epochs for the Warped MS-COCO dataset.For the UDIS-D dataset the training batch size is also set to 16, and the training iteration is 240,000 steps, about 38.4 epochs for the UDIS-D dataset. We have tried a longer training process, but it does not work because the disappearance of the gradient. We set $\omega_1, \omega_2, \omega_3$ to 1,4,16, and the $\lambda_C$ and $\lambda_P$ are 1 and 10, respectively. All the components of this framework are implemented on TensorFlow2~\cite{abadi2016tensorflow}. Both the training and testing are conducted on a single GPU with NVIDIA RTX 3090.
\begin{table*}[ht]
    \centering
    \begin{tabular}{|l||c|c|c|c|c|c|c|c|}
        \hline 
        Dataset&\multicolumn{4}{c|}{Warped MS-COCO}&\multicolumn{4}{c|}{UDIS-D} \\
        \hline
      
        Hardness&  0$\%$-30$\%$ &  30$\%$-60$\%$ & 60$\%$-100$\%$ & Average&0$\%$-30$\%$ &  30$\%$-60$\%$ & 60$\%$-100$\%$  & Average\\
         \hline 
         \hline
        $I_{3\times3}$& 15.5070 & 11.5191 & 8.8709 &  11.6556 & 16.1965 & 13.0621 & 10.8840 &  13.1223 \\
         \hline 
        % SIFT~\cite{lowe2004distinctive}$+$RANSAC~\cite{fischler1981random}& 27.80 & 23.00 & 18.77 &  22.75 & 27.80 & 23.00 & 18.77 &  22.75 \\
        % \hline ORB~\cite{rublee2011orb}$+$RANSAC~\cite{fischler1981random}& 27.80 & 23.00 & 18.77 &  22.75 & 27.80 & 23.00 & 18.77 &  22.75 \\
        \hline 
        \hline 
        UDHN~\cite{nguyen2018unsupervised} & 15.7638 & 11.7356 & 9.0490 &  11.8688& 16.8308 & 13.5883 & 11.2287 &  13.6079 \\
        \hline 
        CA-UDHN~\cite{liu2022content}  & F& F& F& F& F& F& F& F \\
        \hline 
        LB-UDHN~\cite{nie2021unsupervised} &\textcolor{red} {32.9574} & \textcolor{red}{27.9542} & \textcolor{red}{23.5570} &  \textcolor{red}{27.6950} &\textcolor{red}{27.2487} & \textcolor{red}{23.5943} & \textcolor{blue}{20.2476} &  \textcolor{red}{23.3392} \\
        \hline
        \hline
        Ours & \textcolor{blue}{35.3366} & \textcolor{blue}{30.6911} & \textcolor{blue}{26.4958} &  \textcolor{blue}{30.4050} & \textcolor{blue}{27.4885}& \textcolor{blue}{23.7450} & \textcolor{red}{20.1824} &  \textcolor{blue}{23.4280} \\
        \hline
         
    \end{tabular}
    \caption{Comparison of PSNR($\uparrow$) on the overlapping areas with other methods. \textcolor{red}{Red} indicates the second best performance and \textcolor{blue}{Blue} refers to the best result.}
    \label{tab:my_label}
\end{table*}
\begin{table*}[ht]
    \centering
    \begin{tabular}{|l||c|c|c|c|c|c|c|c|}
        \hline 
        Dataset&\multicolumn{4}{c|}{Warped MS-COCO}&\multicolumn{4}{c|}{UDIS-D} \\
        \hline
      
        Hardness&  0$\%$-30$\%$ &  30$\%$-60$\%$ & 60$\%$-100$\%$ & Average&0$\%$-30$\%$ &  30$\%$-60$\%$ & 60$\%$-100$\%$  & Average\\
         \hline 
         \hline
        $I_{3\times3}$& 0.3919 &  0.1665 & 0.0700 &  0.1954 & 0.3831 & 0.1699& 0.0720 &  0.1944 \\
         \hline 
        % SIFT~\cite{lowe2004distinctive}$+$RANSAC~\cite{fischler1981random}& 27.80 & 23.00 & 18.77 &  22.75 & 27.80 & 23.00 & 18.77 &  22.75 \\
        % \hline ORB~\cite{rublee2011orb}$+$RANSAC~\cite{fischler1981random}& 27.80 & 23.00 & 18.77 &  22.75 & 27.80 & 23.00 & 18.77 &  22.75 \\
        \hline 
        \hline 
        UDHN~\cite{nguyen2018unsupervised} & 0.4118 & 0.1875 & 0.0874 &  0.2148 & 0.4115 & 0.1993 & 0.0971 &  0.2217 \\
        \hline 
        CA-UDHN~\cite{liu2022content} & F& F& F& F& F& F& F& F   \\
        \hline 
        LB-UDHN~\cite{nie2021unsupervised} & \textcolor{red}{0.9429} & \textcolor{red}{0.9000} & \textcolor{red}{0.8157} &  \textcolor{red}{0.8790} & \textcolor{red}{0.8875} & \textcolor{red}{0.8117} & \textcolor{red}{0.6569} &  \textcolor{red}{0.7719} \\
        \hline
        \hline
        Ours & \textcolor{blue}{0.9621} & \textcolor{blue}{0.9358} & \textcolor{blue}{0.8778} &  \textcolor{blue}{0.9204} & \textcolor{blue}{0.8949}& \textcolor{blue}{0.8206} & \textcolor{blue}{0.6640} &  \textcolor{blue}{0.7796} \\
        \hline
         
    \end{tabular}
    \caption{Comparison of SSIM($\uparrow$) on the overlapping areas with other methods. \textcolor{red}{Red} indicates the 
 second best performance and \textcolor{blue}{Blue} refers to the  best result.}
    \label{tab:my_label}
\end{table*} 
\subsection{Quantitative Comparison of the Previous Work}
PSNR and SSIM of the overlapping regions are two important factors in homography estimation, which can be calculated as:
 \begin{align} 
     &PSNR= \mathcal{PSNR} (H_{ab}(E)\odot I_a, H_{ab}(I_b)),\\
     &SSIM= \mathcal{SSIM} (H_{ab}(E)\odot I_a, H_{ab}(I_b)),
 \end{align}
 
   Then, we calculate the PSNR and SSIM with the previous work in homography estimation, using all the 10,000 test images in Warped MS-COCO for a comparison. Also, we compare our method in the real dataset, UDIS-D, training a new model, which is fed by the real images. In the real dataset, we also calculate the PSNR and SSIM in 1106 test images, and compare with other methods.

  In the experiment, we classify all the test images into three classes by the hardness of the task. The LB-UDHN~\cite{nie2021unsupervised}method we compared in these two tables is the state-of-the-art method in 8-DOFs homography estimation, and in the synthetic dataset and the real dataset, our method both improves the PSNR  and the SSIM. However, there is a relatively small increase in real dataset, we suppose that our model is large which is not match with the small real dataset, which contains only 10440 training images. 

   Also, in the CA-UDHN~\cite{liu2022content}, we judge it as a failure case, because its result is lower than the $I_{3\times3}$ baseline case. The failure reason for CA-UDHN is that this model is designed for a small-baseline situation without employing a coarse-to-fine strategy.
\subsection{Transfer Learning and Zero Shot Result}
To prove the transfer ability of our large baseline model, we do two parts of experiments: First, we transfer our synthetic pretrained model to downstream real dataset by additionally training 200,000 steps with a 16 batch size. Second, we directly use our synthetic pretrained model to predict the results in real dataset as a zero-shot result.
 \begin{table}[h]
    \centering
    
     \begin{tabular}{|c|c|c|c|c|}
         
        \hline
      
        \multicolumn{2}{|c|}{Method}  &  UDHN~\cite{nguyen2018unsupervised} &LB-UDHN~\cite{nie2021unsupervised}   & Ours  \\

       \hline 
      \multirow{4}*{PSNR} &  0$\%$-30$\%$ & 16.7054 & 27.8380 & \textcolor{blue}{27.8521}   \\
      \cline{2-5}
                     &   30$\%$-60$\%$  & 13.5583 & 23.9915 & \textcolor{blue}{24.0766}  \\
                        \cline{2-5} 
                     &   60$\%$-100$\%$  & 11.2463 & 20.7460 & \textcolor{blue}{20.7750}  \\
                         \cline{2-5}
                     &   Average  & 13.5684 & 23.8363 & \textcolor{blue}{23.8716}  \\
      \hline
      \hline
      \multirow{4}*{SSIM} &  0$\%$-30$\%$ & 0.4015 & 0.8999 & \textcolor{blue}{0.9030}  \\
      \cline{2-5}
                     &   30$\%$-60$\%$  & 0.1957 & 0.8275 & \textcolor{blue}{0.8322} \\
                        \cline{2-5} 
                     &   60$\%$-100$\%$  & 0.1003 & 0.6851 & \textcolor{blue}{0.6878}  \\
                         \cline{2-5}
                     &   Average  & 0.2189 & 0.7917 & \textcolor{blue}{0.7950}  \\
      \hline
 
    \end{tabular}
    \caption{Comparison of PSNR($\uparrow$) and SSIM($\uparrow$)  about the transfer learning results with other methods. \textcolor{blue}{Blue} refers to the  best result.}
    \label{tab:my_label1}
\end{table}

As shown in TABLE \ref{tab:my_label1}, our method outperforms other unsupervised learning methods in transfer learning, which proves the transfer ability of our model. The transfer ability is important for a large model, because with the adaptive ability in other domains, a large model can be widely used in the same task with different scenes.
 \begin{table}[h]
    \centering
    
     \begin{tabular}{|c|c|c|c|c|}
         
        \hline
      
        \multicolumn{2}{|c|}{Method}  &  UDHN~\cite{nguyen2018unsupervised} &LB-UDHN~\cite{nie2021unsupervised}   & Ours  \\

       \hline 
      \multirow{4}*{PSNR} &  0$\%$-30$\%$ & 16.6948 & 26.2967 & \textcolor{blue}{26.8562}   \\
      \cline{2-5}
                     &   30$\%$-60$\%$  & 13.5545 & 22.7501 & \textcolor{blue}{23.2750}  \\
                        \cline{2-5} 
                     &   60$\%$-100$\%$  & 11.2457 & 19.6460 & \textcolor{blue}{19.9778}  \\
                         \cline{2-5}
                     &   Average  & 13.5684 & 22.5590 & \textcolor{blue}{23.0174}  \\
      \hline
      \hline
      \multirow{4}*{SSIM} &  0$\%$-30$\%$ & 0.4006 & 0.8665 & \textcolor{blue}{0.8870}  \\
      \cline{2-5}
                     &   30$\%$-60$\%$  & 0.1951 & 0.7879 & \textcolor{blue}{0.8121} \\
                        \cline{2-5} 
                     &   60$\%$-100$\%$  & 0.1001 & 0.6300 & \textcolor{blue}{0.6568}  \\
                         \cline{2-5}
                     &   Average  & 0.2183 & 0.7477 & \textcolor{blue}{0.7718}  \\
      \hline
 
    \end{tabular}
    \caption{Comparison of PSNR($\uparrow$) and SSIM($\uparrow$)  about the zero shot results with other methods. \textcolor{blue}{Blue} refers to the  best result.}
    \label{tab:my_label2}
\end{table}

As shown in TABLE~\ref{tab:my_label2}, in the zero-shot inference experiment, our method still outperforms the other methods. The zero shot ability reflects the robustness of our model, because with the inference ability, we can directly apply our model in some other real images and get a considerable result.

Then, we compare the whole training process in transfer learning as shown in Fig. 4. In around 240k training steps, there is an obvious increase in our method, however, it is hard to identify an dataset change for LB-UDHN. Our model is more sensitive for the dataset change, which is beneficial for a scene change.
\begin{figure}
	\centering
	\begin{minipage}{0.49\linewidth}
		\centering
		\includegraphics[width=1.0\linewidth]{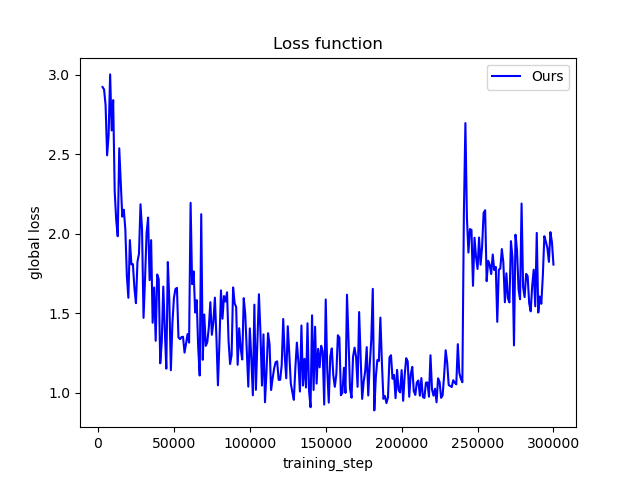}
		  
	\end{minipage}
	%\qquad
	\begin{minipage}{0.49\linewidth}
		\centering
		\includegraphics[width=1.0\linewidth]{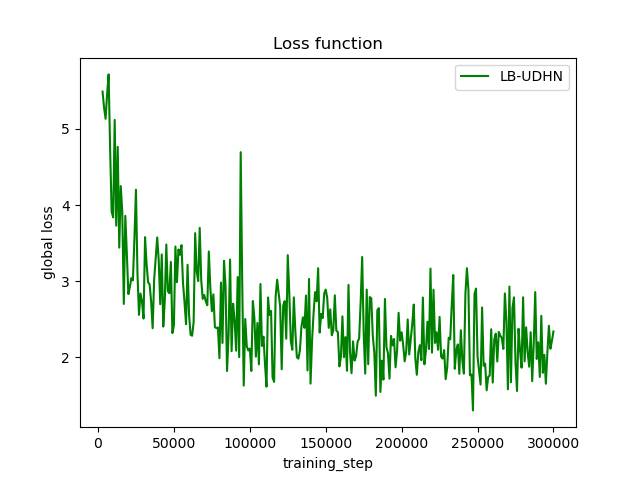}

	\end{minipage}
 \label{transfer} 
 \caption{The comparison of the transfer learning process between ours and LB-UDHN.}
\end{figure}
 
\begin{figure*}
	\centering
        
        \begin{minipage}{0.087\linewidth}
		\centering
		\includegraphics[width=0.58\linewidth]{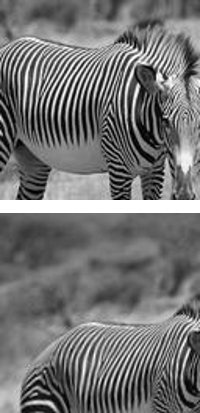}
		 
		\label{chutian1} 
	\end{minipage}
	\begin{minipage}{0.13\linewidth}
		\centering
		\includegraphics[width=0.8\linewidth]{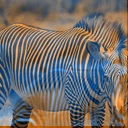}
		 
		\label{chutian1} 
	\end{minipage}
	\begin{minipage}{0.13\linewidth}
		\centering
		\includegraphics[width=0.8\linewidth]{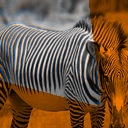}
		 
		\label{chutian2} 
	\end{minipage}
        \begin{minipage}{0.13\linewidth}
		\centering
		\includegraphics[width=0.8\linewidth]{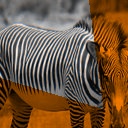}
		 
		\label{chutian1} 
	\end{minipage}
        \begin{minipage}{0.087\linewidth}
		\centering
		\includegraphics[width=0.58\linewidth]{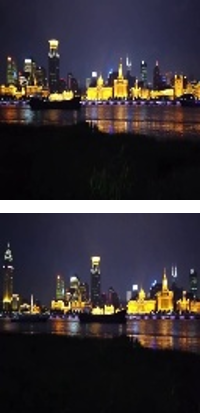}
		 
		\label{chutian1} 
	\end{minipage}
        \begin{minipage}{0.13\linewidth}
		\centering
		\includegraphics[width=0.8\linewidth]{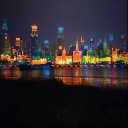}
		 
		\label{chutian1} 
	\end{minipage}
        \begin{minipage}{0.13\linewidth}
		\centering
		\includegraphics[width=0.8\linewidth]{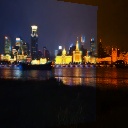}
		 
		\label{chutian1} 
	\end{minipage}
        \begin{minipage}{0.13\linewidth}
		\centering
		\includegraphics[width=0.8\linewidth]{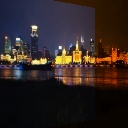}
		 
		\label{chutian1} 
	\end{minipage}
	% \qquad
         
        \vspace{1mm}
        \centering 
	\begin{minipage}{0.087\linewidth}
		\centering
		\includegraphics[width=0.58\linewidth]{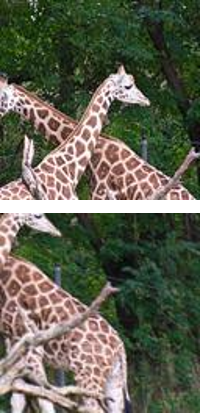}
		 
		\label{chutian1} 
	\end{minipage}
	\begin{minipage}{0.13\linewidth}
		\centering
		\includegraphics[width=0.8\linewidth]{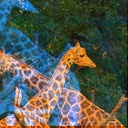}
		 
		\label{chutian1} 
	\end{minipage}
	\begin{minipage}{0.13\linewidth}
		\centering
		\includegraphics[width=0.8\linewidth]{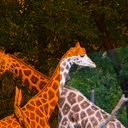}
		 
		\label{chutian2} 
	\end{minipage}
        \begin{minipage}{0.13\linewidth}
		\centering
		\includegraphics[width=0.8\linewidth]{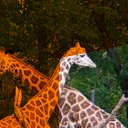}
		 
		\label{chutian1} 
	\end{minipage}
        \begin{minipage}{0.087\linewidth}
		\centering
		\includegraphics[width=0.58\linewidth]{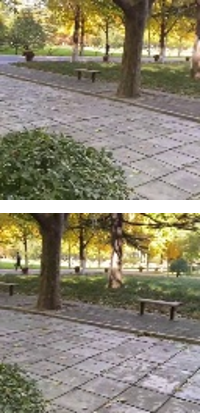}
		 
		\label{chutian1} 
	\end{minipage}
        \begin{minipage}{0.13\linewidth}
		\centering
		\includegraphics[width=0.8\linewidth]{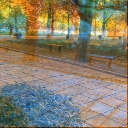}
		 
		\label{chutian1} 
	\end{minipage}
        \begin{minipage}{0.13\linewidth}
		\centering
		\includegraphics[width=0.8\linewidth]{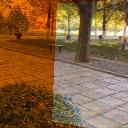}
		 
		\label{chutian1} 
	\end{minipage}
        \begin{minipage}{0.13\linewidth}
		\centering
		\includegraphics[width=0.8\linewidth]{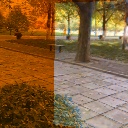}
		 
		\label{chutian1} 
	\end{minipage}
            \\ \hspace*{\fill} \\
        \vspace{1mm}
        \centering 
	\begin{minipage}{0.087\linewidth}
		\centering
		\includegraphics[width=0.58\linewidth]{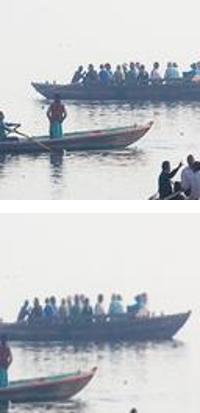}
		 
		\label{chutian1} 
	\end{minipage}
	\begin{minipage}{0.13\linewidth}
		\centering
		\includegraphics[width=0.8\linewidth]{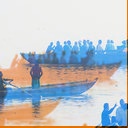}
		 
		\label{chutian1} 
	\end{minipage}
	\begin{minipage}{0.13\linewidth}
		\centering
		\includegraphics[width=0.8\linewidth]{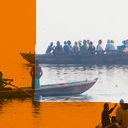}
		 
		\label{chutian2} 
	\end{minipage}
        \begin{minipage}{0.13\linewidth}
		\centering
		\includegraphics[width=0.8\linewidth]{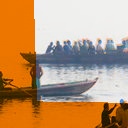}
		 
		\label{chutian1} 
	\end{minipage}
        \begin{minipage}{0.087\linewidth}
		\centering
		\includegraphics[width=0.58\linewidth]{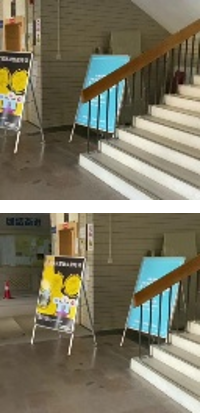}
		 
		\label{chutian1} 
	\end{minipage}
        \begin{minipage}{0.13\linewidth}
		\centering
		\includegraphics[width=0.8\linewidth]{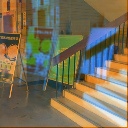}
		 
		\label{chutian1} 
	\end{minipage}
        \begin{minipage}{0.13\linewidth}
		\centering
		\includegraphics[width=0.8\linewidth]{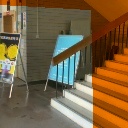}
		 
		\label{chutian1} 
	\end{minipage}
        \begin{minipage}{0.13\linewidth}
		\centering
		\includegraphics[width=0.8\linewidth]{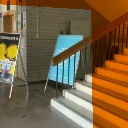}
		 
		\label{chutian1} 
	\end{minipage}
 \\ \hspace*{\fill} \\
         \vspace{1mm}
        \centering 
	\begin{minipage}{0.087\linewidth}
		\centering
		\includegraphics[width=0.58\linewidth]{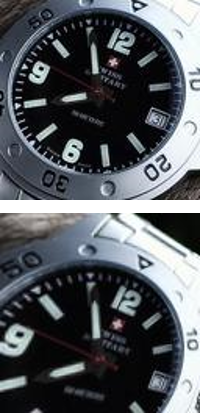}
		 
		\label{chutian1} 
	\end{minipage}
	\begin{minipage}{0.13\linewidth}
		\centering
		\includegraphics[width=0.8\linewidth]{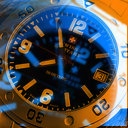}
		 
		\label{chutian1} 
	\end{minipage}
	\begin{minipage}{0.13\linewidth}
		\centering
		\includegraphics[width=0.8\linewidth]{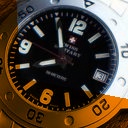}
		 
		\label{chutian2} 
	\end{minipage}
        \begin{minipage}{0.13\linewidth}
		\centering
		\includegraphics[width=0.8\linewidth]{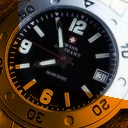}
		 
		\label{chutian1} 
	\end{minipage}
        \begin{minipage}{0.087\linewidth}
		\centering
		\includegraphics[width=0.58\linewidth]{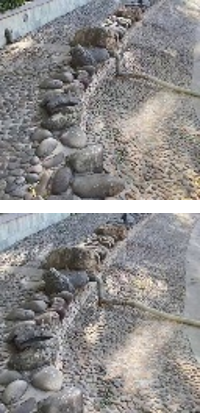}
		 
		\label{chutian1} 
	\end{minipage}
        \begin{minipage}{0.13\linewidth}
		\centering
		\includegraphics[width=0.8\linewidth]{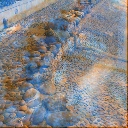}
		 
		\label{chutian1} 
	\end{minipage}
        \begin{minipage}{0.13\linewidth}
		\centering
		\includegraphics[width=0.8\linewidth]{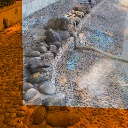}
		 
		\label{chutian1} 
	\end{minipage}
        \begin{minipage}{0.13\linewidth}
		\centering
		\includegraphics[width=0.8\linewidth]{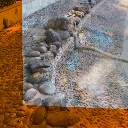}
		 
		\label{chutian1} 
	\end{minipage}
	  \\ \hspace*{\fill} \\
         \vspace{1mm}
        \centering 
	\begin{minipage}{0.087\linewidth}
		\centering
		\includegraphics[width=0.58\linewidth]{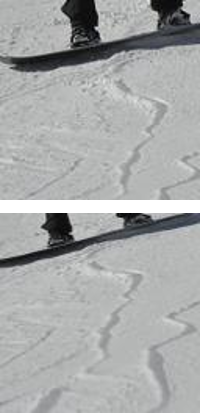}
		 
		\label{chutian1} 
	\end{minipage}
	\begin{minipage}{0.13\linewidth}
		\centering
		\includegraphics[width=0.8\linewidth]{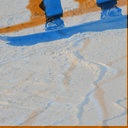}
		 
		\label{chutian1} 
	\end{minipage}
	\begin{minipage}{0.13\linewidth}
		\centering
		\includegraphics[width=0.8\linewidth]{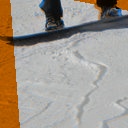}
		 
		\label{chutian2} 
	\end{minipage}
        \begin{minipage}{0.13\linewidth}
		\centering
		\includegraphics[width=0.8\linewidth]{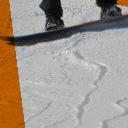}
		 
		\label{chutian1} 
	\end{minipage}
        \begin{minipage}{0.087\linewidth}
		\centering
		\includegraphics[width=0.58\linewidth]{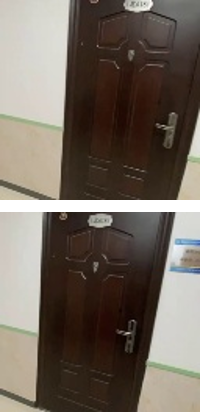}
		 
		\label{chutian1} 
	\end{minipage}
        \begin{minipage}{0.13\linewidth}
		\centering
		\includegraphics[width=0.8\linewidth]{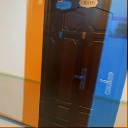}
		 
		\label{chutian1} 
	\end{minipage}
        \begin{minipage}{0.13\linewidth}
		\centering
		\includegraphics[width=0.8\linewidth]{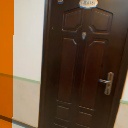}
		 
		\label{chutian1} 
	\end{minipage}
        \begin{minipage}{0.13\linewidth}
		\centering
		\includegraphics[width=0.8\linewidth]{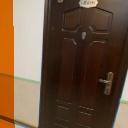}
		 
		\label{chutian1} 
	\end{minipage}
	  \\ \hspace*{\fill} \\
         \vspace{1mm}
        \centering 
	\begin{minipage}{0.087\linewidth}
		\centering
		\includegraphics[width=0.58\linewidth]{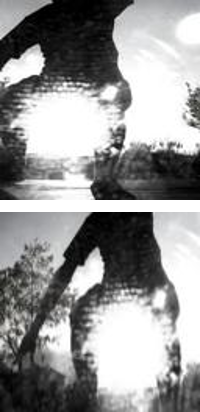}
		 
		\label{chutian1} 
	\end{minipage}
	\begin{minipage}{0.13\linewidth}
		\centering
		\includegraphics[width=0.8\linewidth]{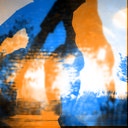}
		 
		\label{chutian1} 
	\end{minipage}
	\begin{minipage}{0.13\linewidth}
		\centering
		\includegraphics[width=0.8\linewidth]{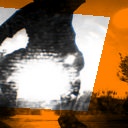}
		 
		\label{chutian2} 
	\end{minipage}
        \begin{minipage}{0.13\linewidth}
		\centering
		\includegraphics[width=0.8\linewidth]{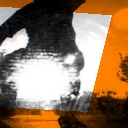}
		 
		\label{chutian1} 
	\end{minipage}
        \begin{minipage}{0.087\linewidth}
		\centering
		\includegraphics[width=0.58\linewidth]{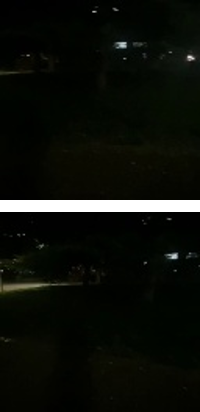}
		 
		\label{chutian1} 
	\end{minipage}
        \begin{minipage}{0.13\linewidth}
		\centering
		\includegraphics[width=0.8\linewidth]{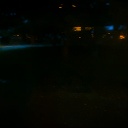}
		 
		\label{chutian1} 
	\end{minipage}
        \begin{minipage}{0.13\linewidth}
		\centering
		\includegraphics[width=0.8\linewidth]{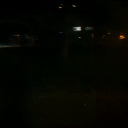}
		 
		\label{chutian1} 
	\end{minipage}
        \begin{minipage}{0.13\linewidth}
		\centering
		\includegraphics[width=0.8\linewidth]{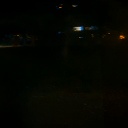}
		 
		\label{chutian1} 
	\end{minipage}
	   \\ \hspace*{\fill} \\
         \vspace{1mm}
        \centering 
	\begin{minipage}{0.087\linewidth}
		\centering
		\includegraphics[width=0.58\linewidth]{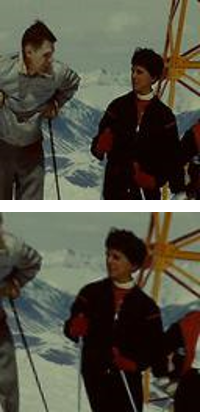}
		 
		\label{chutian1} 
	\end{minipage}
	\begin{minipage}{0.13\linewidth}
		\centering
		\includegraphics[width=0.8\linewidth]{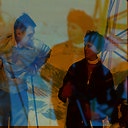}
		 
		\label{chutian1} 
	\end{minipage}
	\begin{minipage}{0.13\linewidth}
		\centering
		\includegraphics[width=0.8\linewidth]{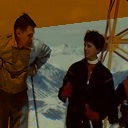}
		 
		\label{chutian2} 
	\end{minipage}
        \begin{minipage}{0.13\linewidth}
		\centering
		\includegraphics[width=0.8\linewidth]{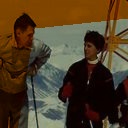}
		 
		\label{chutian1} 
	\end{minipage}
        \begin{minipage}{0.087\linewidth}
		\centering
		\includegraphics[width=0.58\linewidth]{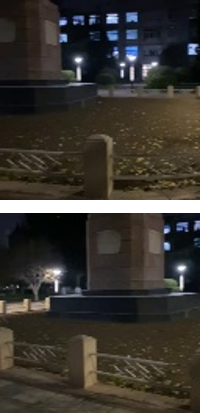}
		 
		\label{chutian1} 
	\end{minipage}
        \begin{minipage}{0.13\linewidth}
		\centering
		\includegraphics[width=0.8\linewidth]{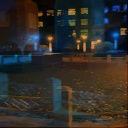}
		 
		\label{chutian1} 
	\end{minipage}
        \begin{minipage}{0.13\linewidth}
		\centering
		\includegraphics[width=0.8\linewidth]{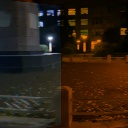}
		 
		\label{chutian1} 
	\end{minipage}
        \begin{minipage}{0.13\linewidth}
		\centering
		\includegraphics[width=0.8\linewidth]{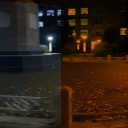}
		 
		\label{chutian1} 
	\end{minipage}
   \\ \hspace*{\fill} \\
         \vspace{1mm}
        \centering 
	\begin{minipage}{0.087\linewidth}
		\centering
		\includegraphics[width=0.58\linewidth]{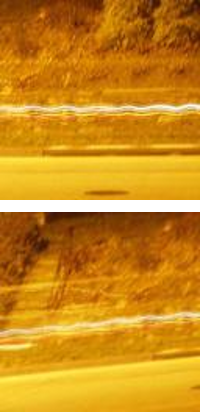}
		 
		\label{chutian1} 
	\end{minipage}
	\begin{minipage}{0.13\linewidth}
		\centering
		\includegraphics[width=0.8\linewidth]{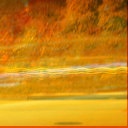}
		 
		\label{chutian1} 
	\end{minipage}
	\begin{minipage}{0.13\linewidth}
		\centering
		\includegraphics[width=0.8\linewidth]{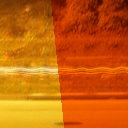}
		 
		\label{chutian2} 
	\end{minipage}
        \begin{minipage}{0.13\linewidth}
		\centering
		\includegraphics[width=0.8\linewidth]{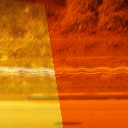}
		 
		\label{chutian1} 
	\end{minipage}
        \begin{minipage}{0.087\linewidth}
		\centering
		\includegraphics[width=0.58\linewidth]{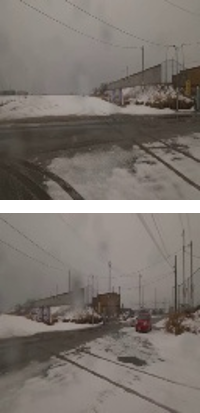}
		 
		\label{chutian1} 
	\end{minipage}
        \begin{minipage}{0.13\linewidth}
		\centering
		\includegraphics[width=0.8\linewidth]{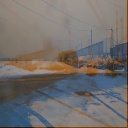}
		 
		\label{chutian1} 
	\end{minipage}
        \begin{minipage}{0.13\linewidth}
		\centering
		\includegraphics[width=0.8\linewidth]{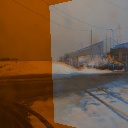}
		 
		\label{chutian1} 
	\end{minipage}
        \begin{minipage}{0.13\linewidth}
		\centering
		\includegraphics[width=0.8\linewidth]{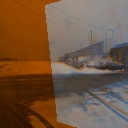}
		 
		\label{chutian1} 
	\end{minipage}
    \\ \hspace*{\fill} \\
         \vspace{1mm}
        \centering 
	\begin{minipage}{0.087\linewidth}
		\centering
		\includegraphics[width=0.58\linewidth]{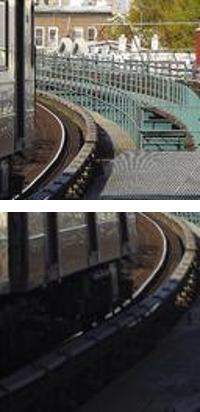}
		 
		\label{chutian1} 
            \centerline{Inputs(s)}
	\end{minipage}
	\begin{minipage}{0.13\linewidth}
		\centering
		\includegraphics[width=0.8\linewidth]{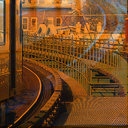}
		 
		\label{chutian1}
             \centerline{Unsupervised(s)}
	\end{minipage}
	\begin{minipage}{0.13\linewidth}
		\centering
		\includegraphics[width=0.8\linewidth]{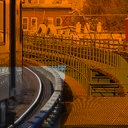}
		 
		\label{chutian2} 
             \centerline{UDHN(s)}
	\end{minipage}
        \begin{minipage}{0.13\linewidth}
		\centering
		\includegraphics[width=0.8\linewidth]{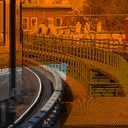}
		 
		\label{chutian1} 
            \centerline{Ours(s)}
	\end{minipage}
        \begin{minipage}{0.087\linewidth}
		\centering
		\includegraphics[width=0.58\linewidth]{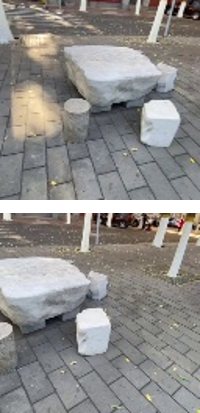}
		 
		\label{chutian1}
             \centerline{Inputs(r)}
	\end{minipage}
        \begin{minipage}{0.13\linewidth}
		\centering
		\includegraphics[width=0.8\linewidth]{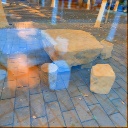}
		 % \centerline{Unsupervised(r)}
		\label{chutian1}
        \centerline{Unsupervised(r)}
	\end{minipage}
        \begin{minipage}{0.13\linewidth}
		\centering
		\includegraphics[width=0.8\linewidth]{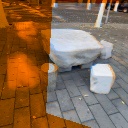}
		   % \centerline{UDHN(r)}
		\label{chutian1} 
  \centerline{UDHN(r)}
	\end{minipage}
        \begin{minipage}{0.13\linewidth}
		\centering
		\includegraphics[width=0.8\linewidth]{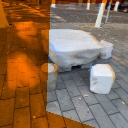}
		   % \centerline{Ours(r)}
		\label{chutian1} 
  \centerline{Ours(r)}
	\end{minipage}
 \caption{The visual quality of homography estimation. After combining the source and warped images by different channel settings, there is no blur, ghost effect in our method. s is the result in synthetic dataset. r is the result in real dataset.  }
 \label{fusion}
\end{figure*}
\subsection{Visual Quality}
One of the important applications for homography estimation is image stitching. By observing the visual effect of the overlapping area, we can estimate the matching degree of the two result images.

In this part, as show in Fig. ~\ref{fusion}, we visualize our quality effect by combining the result images in one plane. Setting different channels for the source images and warped images, we can more clearly observe the effect. In our method, we successfully avoid the blur and ghost effect, and match the margin areas.
 \subsection{Ablation Study and Discussion}
 In this part, we do ablation experiment for the transformer backbone and channel attention. The transformer backbone ablation experiment is about the comparison between transformer-style extractor the cnn-style extractor. The experiment about channel attention is that we drop the channel attention layer for the correlation volume and observe the result.

 As shown in TABLE ~\ref{tbl:table1}, we test our result in the synthetic dataset. Our performance on the testing dataset both reduce after dropping the channel attention or replacing the transformer backbone with cnn backbone.
 \begin{figure*}
     \centering
     \includegraphics[width=1\textwidth]{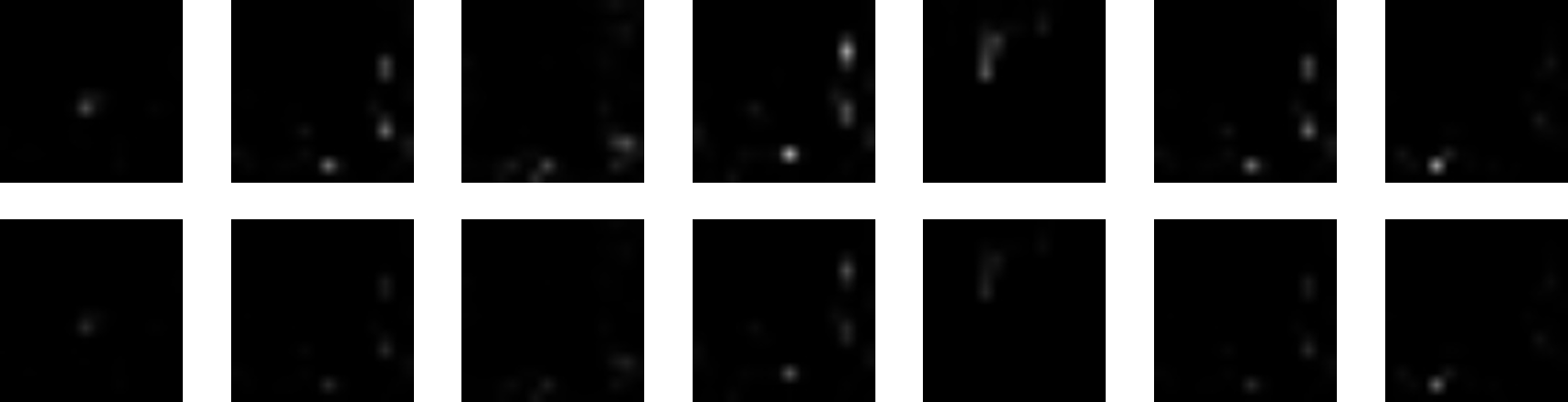} 
     \caption{The comparison of the single channel map for channel attention experiment. The first line is the channel map without channel attention. The second line is the same number channel map with channel attention. There is an obvious rejection for some weak correlation in the correlation volume shown in the second line.}
     \label{channel}
 \end{figure*}
 \begin{table}[H]
\centering
\caption{Ablation study for transformer extractor and channel attention}
\begin{tabular}{llll} 
\toprule
  \multicolumn{2}{l}{Ablation Structure }   \\
\cline{1-2}   
      Transformer Extractor& Channel Attention &  \  SSIM & PSNR  \\
  \midrule
      $\times$ & \checkmark   & 0.9139 & 29.6863   \\
   \checkmark & $\times$  & 0.9156 &  30.0380   \\
   \checkmark & \checkmark  & \textcolor{blue}{0.9204 }  &  \textcolor{blue}{30.4050 }    \\
   
  \bottomrule
  \end{tabular}
  \label{tbl:table1}
\end{table}
 \textbf{CNN and Transformer Comparison} First, we visualize the feature maps extracted by the cnn and transformer comparison experiments. As shown in Fig. ~\ref{cnn}, our trained performance better represent the basical feature information from the original input images compared with cnn methods. In the homography task, some concrete feature representations are better than the abstract feature representations for the following position searching. Therefore, after the training process, the transformer-style extractor outperforms the cnn-style extractor for feature map extracting in the homography estimation task.

  \begin{figure}[H]
     \centering
     \includegraphics[width=0.35\textwidth]{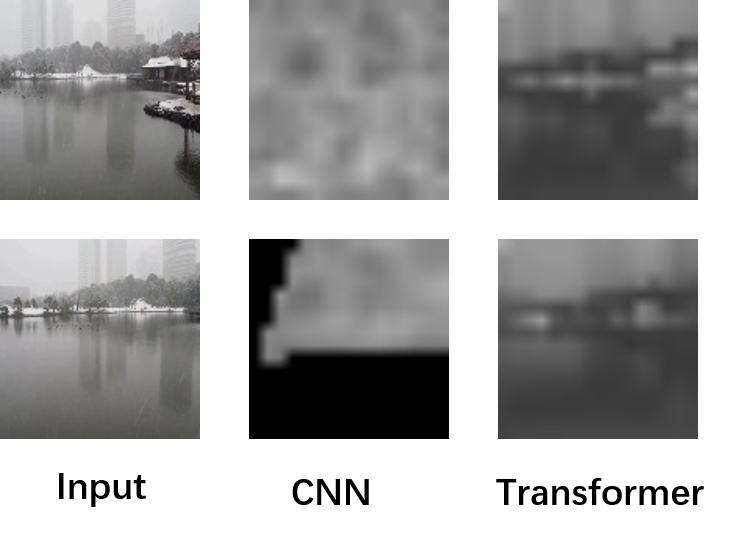}
     \caption{The different feature maps extracted from cnn and transformer.}
     \label{cnn}
 \end{figure}
\begin{figure}[H]
     \centering
     \includegraphics[width=0.5\textwidth]{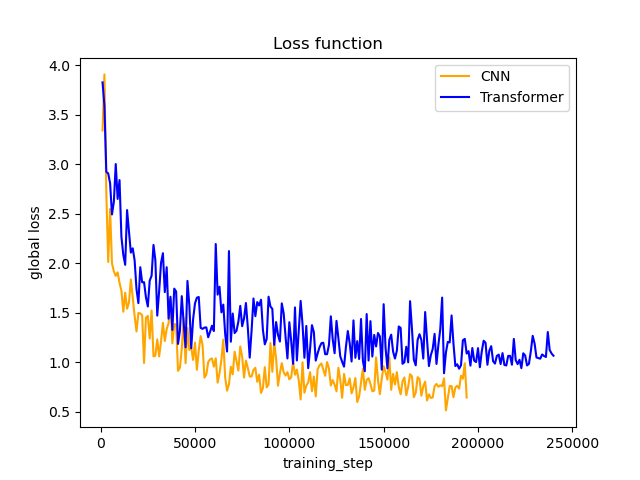}
     \caption{The whole training process comparison between cnn and transformer. After 200k steps, the cnn loss outputs a nan for a gradient extinction. }
     \label{training}
 \end{figure}

  After that, we compare the whole training process between cnn and transformer as shown in Fig. ~\ref{training}. We set nearly the same traing step for a comparison. However, after 200k steps, the cnn layer output a nan for the global loss because of the Gradient extinction. In contrast, the transformer-style extractor still shows a descent for the global loss in the training process. Thus, our model is more potential for a large-scale training compared with traditional cnn methods. Compared with cnn, transformer has more information interaction because of a larger range modelling, so it supports more training steps to obtain a global information interaction.  
 \textbf{Channel Attention Ablation Study} In this part, we visualize the single channel map, between with channel attention model and no channel attention model as shown in Fig. ~\ref{channel}. Every single channel map stores the position information of which pixel in the other input images is a correlation with the pixel represented by the channel map. In the first line, there are several correlation candidate points. After the channel attention, in the second line, some weak correlation points are rejected. Because in the last step, we need input the correlation volume after channel attention to a regression layer, the channel attention can alleviate the burden of the regressing process without too much search for the matched points. Also, it will avoid some mismatch search to some extent, by filtering some weak correlation points.
\section{Future Work}
 Because of the resolution limitation of swin-transformer model, our model cannot extend to arbitrary resolution situation. In the future work, there may be some exploration for the interpolation method for the position embedding in this model to develop a large baseline model for arbitrary resolution.

The other part of work can be explored is that to develop a multi-grid method for this model, as the method developed in ~\cite{nie2021depth}. Combining  transformer large-scale training and accurate homography method, the performance may improve and tranfer to more scenes and tasks.
\section{Conclusion}
In this paper, we propose an all-attention layer based homography estimation network. In the feature extraction stage, we employ both CNN layer and Swin Transformer block, for shallow feature extraction and deep feature extraction, respectively. Then, we design a cross non-local attention mechanism, which searches a match region between the feature maps and aligns the feature maps in advance. For the homography regression layer, before predicting the homography transformation, a channel attenton layer is adopted for the correlation volume, which drops out some outlier points by decreasing the attention weights. In the experiment section, we compare our method with the state-of-the-art previous work, we outperforms their work in a large range test dataset. Also, we transfer our pretrained model to real dataset, the experiment results indicate the generalization property of our proposed baseline model.


\begin{thebibliography}{10}
\providecommand{\url}[1]{#1}
\csname url@samestyle\endcsname
\providecommand{\newblock}{\relax}
\providecommand{\bibinfo}[2]{#2}
\providecommand{\BIBentrySTDinterwordspacing}{\spaceskip=0pt\relax}
\providecommand{\BIBentryALTinterwordstretchfactor}{4}
\providecommand{\BIBentryALTinterwordspacing}{\spaceskip=\fontdimen2\font plus
\BIBentryALTinterwordstretchfactor\fontdimen3\font minus
  \fontdimen4\font\relax}
\providecommand{\BIBforeignlanguage}[2]{{%
\expandafter\ifx\csname l@#1\endcsname\relax
\typeout{** WARNING: IEEEtran.bst: No hyphenation pattern has been}%
\typeout{** loaded for the language `#1'. Using the pattern for}%
\typeout{** the default language instead.}%
\else
\language=\csname l@#1\endcsname
\fi
#2}}
\providecommand{\BIBdecl}{\relax}
\BIBdecl

\bibitem{zhang2022image}
Z.~Zhang, X.~Yu, and X.~Yang, ``Image alignment using mixture models for
  discontinuous deformations,'' \emph{Signal Processing}, vol. 195, p. 108467,
  2022.

\bibitem{zhang2021natural}
Z.~Zhang, X.~Yang, and C.~Xu, ``Natural image stitching with layered warping
  constraint,'' \emph{IEEE Transactions on Multimedia}, 2021.

\bibitem{detone2016deep}
D.~DeTone, T.~Malisiewicz, and A.~Rabinovich, ``Deep image homography
  estimation,'' \emph{arXiv preprint arXiv:1606.03798}, 2016.

\bibitem{nguyen2018unsupervised}
T.~Nguyen, S.~W. Chen, S.~S. Shivakumar, C.~J. Taylor, and V.~Kumar,
  ``Unsupervised deep homography: A fast and robust homography estimation
  model,'' \emph{IEEE Robotics and Automation Letters}, vol.~3, no.~3, pp.
  2346--2353, 2018.

\bibitem{zhang2020content}
J.~Zhang, C.~Wang, S.~Liu, L.~Jia, N.~Ye, J.~Wang, J.~Zhou, and J.~Sun,
  ``Content-aware unsupervised deep homography estimation,'' in \emph{European
  Conference on Computer Vision}.\hskip 1em plus 0.5em minus 0.4em\relax
  Springer, 2020, pp. 653--669.

\bibitem{nie2020view}
L.~Nie, C.~Lin, K.~Liao, M.~Liu, and Y.~Zhao, ``A view-free image stitching
  network based on global homography,'' \emph{Journal of Visual Communication
  and Image Representation}, vol.~73, p. 102950, 2020.

\bibitem{liu2021swin}
Z.~Liu, Y.~Lin, Y.~Cao, H.~Hu, Y.~Wei, Z.~Zhang, S.~Lin, and B.~Guo, ``Swin
  transformer: Hierarchical vision transformer using shifted windows,'' in
  \emph{Proceedings of the IEEE/CVF International Conference on Computer
  Vision}, 2021, pp. 10\,012--10\,022.

\bibitem{carion2020end}
N.~Carion, F.~Massa, G.~Synnaeve, N.~Usunier, A.~Kirillov, and S.~Zagoruyko,
  ``End-to-end object detection with transformers,'' in \emph{European
  conference on computer vision}.\hskip 1em plus 0.5em minus 0.4em\relax
  Springer, 2020, pp. 213--229.

\bibitem{dosovitskiy2020image}
A.~Dosovitskiy, L.~Beyer, A.~Kolesnikov, D.~Weissenborn, X.~Zhai,
  T.~Unterthiner, M.~Dehghani, M.~Minderer, G.~Heigold, S.~Gelly \emph{et~al.},
  ``An image is worth 16x16 words: Transformers for image recognition at
  scale,'' \emph{arXiv preprint arXiv:2010.11929}, 2020.

\bibitem{liang2021swinir}
J.~Liang, J.~Cao, G.~Sun, K.~Zhang, L.~Van~Gool, and R.~Timofte, ``Swinir:
  Image restoration using swin transformer,'' in \emph{Proceedings of the
  IEEE/CVF International Conference on Computer Vision}, 2021, pp. 1833--1844.

\bibitem{vaswani2017attention}
A.~Vaswani, N.~Shazeer, N.~Parmar, J.~Uszkoreit, L.~Jones, A.~N. Gomez,
  {\L}.~Kaiser, and I.~Polosukhin, ``Attention is all you need,''
  \emph{Advances in neural information processing systems}, vol.~30, 2017.

\bibitem{fischler1981random}
M.~A. Fischler and R.~C. Bolles, ``Random sample consensus: a paradigm for
  model fitting with applications to image analysis and automated
  cartography,'' \emph{Communications of the ACM}, vol.~24, no.~6, pp.
  381--395, 1981.

\bibitem{hu2018squeeze}
J.~Hu, L.~Shen, and G.~Sun, ``Squeeze-and-excitation networks,'' in
  \emph{Proceedings of the IEEE conference on computer vision and pattern
  recognition}, 2018, pp. 7132--7141.

\bibitem{woo2018cbam}
S.~Woo, J.~Park, J.-Y. Lee, and I.~S. Kweon, ``Cbam: Convolutional block
  attention module,'' in \emph{Proceedings of the European conference on
  computer vision (ECCV)}, 2018, pp. 3--19.

\bibitem{nie2021depth}
L.~Nie, C.~Lin, K.~Liao, S.~Liu, and Y.~Zhao, ``Depth-aware multi-grid deep
  homography estimation with contextual correlation,'' \emph{arXiv preprint
  arXiv:2107.02524}, 2021.

\bibitem{nie2021unsupervised}
------, ``Unsupervised deep image stitching: Reconstructing stitched features
  to images,'' \emph{IEEE Transactions on Image Processing}, vol.~30, pp.
  6184--6197, 2021.

\bibitem{ye2021motion}
N.~Ye, C.~Wang, H.~Fan, and S.~Liu, ``Motion basis learning for unsupervised
  deep homography estimation with subspace projection,'' in \emph{Proceedings
  of the IEEE/CVF International Conference on Computer Vision}, 2021, pp.
  13\,117--13\,125.

\bibitem{hong2022unsupervised}
M.~Hong, Y.~Lu, N.~Ye, C.~Lin, Q.~Zhao, and S.~Liu, ``Unsupervised homography
  estimation with coplanarity-aware gan,'' in \emph{Proceedings of the IEEE/CVF
  Conference on Computer Vision and Pattern Recognition}, 2022, pp.
  17\,663--17\,672.

\bibitem{xiao2021early}
T.~Xiao, M.~Singh, E.~Mintun, T.~Darrell, P.~Doll{\'a}r, and R.~Girshick,
  ``Early convolutions help transformers see better,'' \emph{Advances in Neural
  Information Processing Systems}, vol.~34, pp. 30\,392--30\,400, 2021.

\bibitem{he2016deep}
K.~He, X.~Zhang, S.~Ren, and J.~Sun, ``Deep residual learning for image
  recognition,'' in \emph{Proceedings of the IEEE conference on computer vision
  and pattern recognition}, 2016, pp. 770--778.

\bibitem{ba2016layer}
J.~L. Ba, J.~R. Kiros, and G.~E. Hinton, ``Layer normalization,'' \emph{arXiv
  preprint arXiv:1607.06450}, 2016.

\bibitem{sun2018pwc}
D.~Sun, X.~Yang, M.-Y. Liu, and J.~Kautz, ``Pwc-net: Cnns for optical flow
  using pyramid, warping, and cost volume,'' in \emph{Proceedings of the IEEE
  conference on computer vision and pattern recognition}, 2018, pp. 8934--8943.

\bibitem{teed2020raft}
Z.~Teed and J.~Deng, ``Raft: Recurrent all-pairs field transforms for optical
  flow,'' in \emph{European conference on computer vision}.\hskip 1em plus
  0.5em minus 0.4em\relax Springer, 2020, pp. 402--419.

\bibitem{lin2014microsoft}
T.-Y. Lin, M.~Maire, S.~Belongie, J.~Hays, P.~Perona, D.~Ramanan,
  P.~Doll{\'a}r, and C.~L. Zitnick, ``Microsoft coco: Common objects in
  context,'' in \emph{European conference on computer vision}.\hskip 1em plus
  0.5em minus 0.4em\relax Springer, 2014, pp. 740--755.

\bibitem{kingma2014adam}
D.~P. Kingma and J.~Ba, ``Adam: A method for stochastic optimization,''
  \emph{arXiv preprint arXiv:1412.6980}, 2014.

\bibitem{rublee2011orb}
E.~Rublee, V.~Rabaud, K.~Konolige, and G.~Bradski, ``Orb: An efficient
  alternative to sift or surf,'' in \emph{2011 International conference on
  computer vision}.\hskip 1em plus 0.5em minus 0.4em\relax Ieee, 2011, pp.
  2564--2571.

\bibitem{bay2006surf}
H.~Bay, T.~Tuytelaars, and L.~Van~Gool, ``Surf: Speeded up robust features,''
  in \emph{Computer Vision--ECCV 2006: 9th European Conference on Computer
  Vision, Graz, Austria, May 7-13, 2006. Proceedings, Part I 9}.\hskip 1em plus
  0.5em minus 0.4em\relax Springer, 2006, pp. 404--417.

\bibitem{lowe2004distinctive}
D.~G. Lowe, ``Distinctive image features from scale-invariant keypoints,''
  \emph{International journal of computer vision}, vol.~60, no.~2, pp. 91--110,
  2004.

\bibitem{mur2015orb}
R.~Mur-Artal, J.~M.~M. Montiel, and J.~D. Tardos, ``Orb-slam: a versatile and
  accurate monocular slam system,'' \emph{IEEE transactions on robotics},
  vol.~31, no.~5, pp. 1147--1163, 2015.

\bibitem{gao2011constructing}
J.~Gao, S.~J. Kim, and M.~S. Brown, ``Constructing image panoramas using
  dual-homography warping,'' in \emph{CVPR 2011}.\hskip 1em plus 0.5em minus
  0.4em\relax IEEE, 2011, pp. 49--56.

\bibitem{hartley2003multiple}
R.~Hartley and A.~Zisserman, \emph{Multiple view geometry in computer
  vision}.\hskip 1em plus 0.5em minus 0.4em\relax Cambridge university press,
  2003.

\bibitem{zaragoza2013projective}
J.~Zaragoza, T.-J. Chin, M.~S. Brown, and D.~Suter, ``As-projective-as-possible
  image stitching with moving dlt,'' in \emph{Proceedings of the IEEE
  conference on computer vision and pattern recognition}, 2013, pp. 2339--2346.

\bibitem{brown2007automatic}
M.~Brown and D.~G. Lowe, ``Automatic panoramic image stitching using invariant
  features,'' \emph{International journal of computer vision}, vol.~74, no.~1,
  pp. 59--73, 2007.

\bibitem{chang2017clkn}
C.-H. Chang, C.-N. Chou, and E.~Y. Chang, ``Clkn: Cascaded lucas-kanade
  networks for image alignment,'' in \emph{Proceedings of the IEEE conference
  on computer vision and pattern recognition}, 2017, pp. 2213--2221.

\bibitem{erlik2017homography}
F.~Erlik~Nowruzi, R.~Laganiere, and N.~Japkowicz, ``Homography estimation from
  image pairs with hierarchical convolutional networks,'' in \emph{Proceedings
  of the IEEE international conference on computer vision workshops}, 2017, pp.
  913--920.

\bibitem{nie2020learning}
L.~Nie, C.~Lin, K.~Liao, and Y.~Zhao, ``Learning edge-preserved image stitching
  from large-baseline deep homography,'' \emph{arXiv preprint
  arXiv:2012.06194}, 2020.

\bibitem{liu2022content}
S.~Liu, N.~Ye, C.~Wang, K.~Luo, J.~Wang, and J.~Sun, ``Content-aware
  unsupervised deep homography estimation and beyond,'' \emph{IEEE Transactions
  on Pattern Analysis and Machine Intelligence}, 2022.

\bibitem{liu2022unsupervised}
S.~Liu, Y.~Lu, H.~Jiang, N.~Ye, C.~Wang, and B.~Zeng, ``Unsupervised global and
  local homography estimation with motion basis learning,'' \emph{IEEE
  Transactions on Pattern Analysis and Machine Intelligence}, 2022.

\bibitem{koguciuk2021perceptual}
D.~Koguciuk, E.~Arani, and B.~Zonooz, ``Perceptual loss for robust unsupervised
  homography estimation,'' in \emph{Proceedings of the IEEE/CVF Conference on
  Computer Vision and Pattern Recognition}, 2021, pp. 4274--4283.

\bibitem{gatys2016image}
L.~A. Gatys, A.~S. Ecker, and M.~Bethge, ``Image style transfer using
  convolutional neural networks,'' in \emph{Proceedings of the IEEE conference
  on computer vision and pattern recognition}, 2016, pp. 2414--2423.

\bibitem{Ye_2021_ICCV}
N.~Ye, C.~Wang, H.~Fan, and S.~Liu, ``Motion basis learning for unsupervised
  deep homography estimation with subspace projection,'' in \emph{Proceedings
  of the IEEE/CVF International Conference on Computer Vision (ICCV)}, October
  2021, pp. 13\,117--13\,125.

\bibitem{medsker2001recurrent}
L.~R. Medsker and L.~Jain, ``Recurrent neural networks,'' \emph{Design and
  Applications}, vol.~5, pp. 64--67, 2001.

\bibitem{hochreiter1997long}
S.~Hochreiter and J.~Schmidhuber, ``Long short-term memory,'' \emph{Neural
  computation}, vol.~9, no.~8, pp. 1735--1780, 1997.

\bibitem{fu2019dual}
J.~Fu, J.~Liu, H.~Tian, Y.~Li, Y.~Bao, Z.~Fang, and H.~Lu, ``Dual attention
  network for scene segmentation,'' in \emph{Proceedings of the IEEE/CVF
  conference on computer vision and pattern recognition}, 2019, pp. 3146--3154.

\bibitem{wang2018non}
X.~Wang, R.~Girshick, A.~Gupta, and K.~He, ``Non-local neural networks,'' in
  \emph{Proceedings of the IEEE conference on computer vision and pattern
  recognition}, 2018, pp. 7794--7803.

\bibitem{arnab2021vivit}
A.~Arnab, M.~Dehghani, G.~Heigold, C.~Sun, M.~Lu{\v{c}}i{\'c}, and C.~Schmid,
  ``Vivit: A video vision transformer,'' in \emph{Proceedings of the IEEE/CVF
  International Conference on Computer Vision}, 2021, pp. 6836--6846.

\bibitem{bain2021frozen}
M.~Bain, A.~Nagrani, G.~Varol, and A.~Zisserman, ``Frozen in time: A joint
  video and image encoder for end-to-end retrieval,'' in \emph{Proceedings of
  the IEEE/CVF International Conference on Computer Vision}, 2021, pp.
  1728--1738.

\bibitem{liu2022video}
Z.~Liu, J.~Ning, Y.~Cao, Y.~Wei, Z.~Zhang, S.~Lin, and H.~Hu, ``Video swin
  transformer,'' in \emph{Proceedings of the IEEE/CVF Conference on Computer
  Vision and Pattern Recognition}, 2022, pp. 3202--3211.

\bibitem{lin2022ds}
A.~Lin, B.~Chen, J.~Xu, Z.~Zhang, G.~Lu, and D.~Zhang, ``Ds-transunet: Dual
  swin transformer u-net for medical image segmentation,'' \emph{IEEE
  Transactions on Instrumentation and Measurement}, 2022.

\bibitem{abadi2016tensorflow}
M.~Abadi, P.~Barham, J.~Chen, Z.~Chen, A.~Davis, J.~Dean, M.~Devin,
  S.~Ghemawat, G.~Irving, M.~Isard \emph{et~al.}, ``Tensorflow: a system for
  large-scale machine learning.'' in \emph{Osdi}, vol.~16, no. 2016.\hskip 1em
  plus 0.5em minus 0.4em\relax Savannah, GA, USA, 2016, pp. 265--283.

\end{thebibliography}
\end{document}